\documentclass[10pt, a4paper]{article}
\usepackage{lrec}
\usepackage{graphicx}
\usepackage{tabularx}
\usepackage{soul}
\usepackage{epstopdf}
\usepackage[latin1]{inputenc}
\usepackage{hyperref}
\usepackage{xstring}
\usepackage{listings}
\usepackage{caption, newfloat}
\usepackage{pgfplots}
\pgfplotsset{compat=newest}
\usepackage{multirow}

\newcommand{\eat}[1]{}

\def\2ps{{{2\textit{ps}}}}

\newcommand{\mention}[1]{\underline{{\color{blue}\textsl{{\color{blue} #1}}}}}
\newcommand{\term}[1]{{\color{blue}\textsl{{\color{blue} ``#1''}}}}
\newcommand{\entity}[1]{{\color{blue}\textsl{{\color{blue} uri:#1}}}}

\def\Wiki{\mbox{\texttt{Wiki}}}
\def\Trans{\mbox{\texttt{Trans}}}
\def\ASR{\mbox{\texttt{ASR}}}

\newcommand{\framedbox}[2][0.96\textwidth]{
  \centering
  \tikzstyle{mybox} = [draw=black,line width=1pt,inner sep=4pt]
  \begin{tikzpicture}
   \node [mybox] (box){%
    \begin{minipage}{#1}{#2}\end{minipage}
   };
  \end{tikzpicture}
}
 
\DeclareFloatingEnvironment[listname=Text,name=Text]{story}
\newcommand{\secref}[1]{\StrSubstitute{\getrefnumber{#1}}{.}{ }}

\title{What did you Mention? A Large Scale Mention Detection Benchmark for Spoken and Written Text*
}

\name{Yosi Mass, Lili Kotlerman, Shachar Mirkin,  Elad Venezian, Gera Witzling, Noam Slonim}
\address{IBM Haifa Reseacrh Lab \\
         Haifa University, Mount Carmel, Haifa, HA 31905, Israel \\
         \{yosimass,lili.kotlerman,shacharm,eladv,geraw,noams\}@il.ibm.com\\}

\abstract{
\thanks{* This work is licensed under a CC BY-ND license}
We describe a large, high-quality benchmark for the evaluation of Mention Detection tools. 
The benchmark contains annotations of both named entities as well as other types of entities, 
annotated on different types of text,
ranging from clean text taken from Wikipedia, to noisy spoken data.
The benchmark was built through a highly controlled crowd sourcing process to ensure its quality.
We describe the benchmark, the process and the guidelines that were used to build it. We then demonstrate the results of a state-of-the-art system running on that benchmark.  
\vspace{0.1in}\newline   
}

\begin{document}

\maketitleabstract

\section{Introduction}
\label{sec:introduction}
Extracting semantic information from text is a fundamental task in various NLP (Natural Language Processing) applications such as Information 
Retrieval, Question Answering, 
% ns Argument Construction, Compliance and more. 
Text Similarity, Argument Construction, and more. 
The task, referred to as Mention Detection (or Entity Linking or Wikification) links terms 
% ns (a term can be one or more tokens) 
(i.e., a single token, or a consecutive sequence of tokens)
from unstructured text to nodes (i.e., entities) in a Knowledge Base, such as 
e.g., DBPedia\footnote{http://dbpedia.org}.
A Mention is thus a tuple $(t,s,u)$ where $t$ is a term that appears in span\footnote{A span is the begin and end offsets of the term in the text} $s$ in the given text, and $u$ is an entity in the Knowledge Base. 

In general, entities in a Knowledge Base can be divided into two main types. 
% ns Named entities such as persons, organizations and locations and all other entities. 
Well-defined {\it named entities\/} such as persons, organizations, and locations; and all other entities. 
% ns Consider for example the annotated text below

Consider for example the annotated Text~\ref{box:box1} below. 
The term \term{Attorney General Edwin Meese} is mapped to the person entity \entity{Edwin\_Meese}, 
while \term{proximity} is mapped to the general entity \entity{Distance}. 

\begin{story}[ht]
  \framedbox[0.4\textwidth]
		{\mention{Attorney General Edwin Meese} determined that the \mention{headquarters} had to be located in close \mention{proximity} to the \mention{Attorney General's office. }}
		\caption[text]{}
  \label{box:box1}
\end{story}

%\noindent 
%\fbox{
			%\parbox{1.0\linewidth} {%
							 %However, then-\mention{Attorney General Edwin Meese} determined that the \mention{headquarters} had to be located in close \mention{proximity} to \ldots%
							%}%
	%}	

% ns Most of existing Mention detection tools focus on extracting named entities only since that task is well defined. 
Most of the existing Mention Detection tools focus on extracting named entities, probably since this task is more 
easily defined. 
The task of linking all types of entities is more vague~\cite{TACL15} and requires clear guidelines on what to annotate, 
how to deal with nested terms, and how to resolve specificity of entities.      

However, extracting all types of entities, not only named entities,  is crucial for semantic understanding of text. Consider for example the task of semantic similarity. It is easy to see that Text~\ref{box:box1} and Text~\ref{box:box11} are quite similar though the wordings are quite different. Such a similarity can be inferred thanks to the mapping of \mention{proximity} and \mention{distance} to the same general entity and the mapping of \mention{headquarters} and \mention{base of operation} to the same general entity. 

\begin{story}[ht]
  \framedbox[0.4\textwidth]
		{\mention{Attorney General Edwin Meese} determined that the \mention{base of operation} was located in close \mention{distance} to the \mention{Attorney General's office. }}
		\caption[text]{}
  \label{box:box11}
\end{story}

We illustrate now the difficulties in building a benchmark for all types of entities. 
Regarding the issue of what to annotate, consider again Text~\ref{box:box1}. 
% ns Does it make sense to create mentions for very general terms like (\term{that},\entity{That})? 
Shall one link the general term \term{that} to \entity{That}? 
% ns And what about the mention (\term{determined},\entity{Determinacy}). 
% ns Do they contribute anything to the semantic understanding of the text?
% ns Do they contribute anything to the semantic understanding of the text?
Similarly, shall one link \term{determined} to \entity{Determinacy} and will this
contribute anything to the semantic understanding of the text?

Another issue 
to consider % ns
is the nesting of Mentions. 
In Text~\ref{box:box1}, should 
% there be a mention (\term{office},entity{Office}) nested inside
we link \term{office} to \entity{Office} even though it is nested within the Mention 
(\term{Attorney General's office}, \entity{United\_States\_Attorney\_General})?

\begin{story}[ht]
  \framedbox[0.4\textwidth]
		{The~\mention{empire} ended in $1889$, when~\mention{Pedro II} was~\mention{deposed} \ldots}
		\caption[text]{}
  \label{box:box2}
\end{story}

%\noindent 
%\fbox{
			%\parbox{1.0\linewidth} {%
					%The~\mention{empire} ended in $1889$, when~\mention{Pedro II} was~\mention{deposed} by a\\~\mention{military coup} %
					%}%
			%}
			%

The last issue with general entities is the specificity of Mentions. Consider Text~\ref{box:box2}. 
Should the term \term{empire} be linked to the general entity \entity{Empire} or to the more 
specific entity \entity{Empire\_of\_Brazil}? 

Thus, % ns Consequently, most of existing Mention Detection benchmarks contain named entities only. 
building a benchmark for the evaluation of Mention Detection tools for all types of entities, requires 
carefully % ns a careful 
crafted guidelines. 
Moreover, all existing benchmarks annotate relatively clean 
and well--phrased % ns
text taken from Wikipedia or from newspapers. 
% There is no Mention Detection benchmark for spoken data. Such a data is relatively noisy 
To the best of our knowledge there is no Mention Detection benchmark for spoken data, which naturally is more noisy
and thus poses new challenges for Mention Detection.

Furthermore, NLP applications would work better if they can exploit Mentions that are found all over the given texts. Wikipedia for example, contains hyperlinks only in the first appearance of a Mention while other appearances in a document are not hyperlinked. The current benchmark contains a full coverage of Mentions all over the annotated texts and 
thus it enables evaluation of Mention Detection tools that require such property.

In this paper we present 
% ns a benchmark that covers both named entities as well as general entities, and both written, clean data, as well as noisy, spoken data.
a comprehensive benchmark data covering both named entities as well as general entities, for both written--text data
as well as noisy, spoken data.

% ns
% The benchmark consists of two diversified datasets with different characteristics. 
% The first dataset contains sentences from Wikipedia. The second 
% contains sentences from spoken data~\cite{DEBATER-OSA}.

Each dataset contains $1000$ sentences that were annotated through a 
carefully % ns 
controlled crowd sourcing process. 
% ns Each sentence was labeled by $10$ labelers in two rounds of detection and confirmation, 
Each sentence was labeled by two rounds of {\it detection\/} and {\it confirmation\/}, done by $10$ labelers each,
resulting in about $6500$ Mentions in each of the two datasets. 
% ns To the best of our knowledge this is the first benchmark on spoken data. 
We describe the process that was used to build the benchmark and 
further % ns then 
present a simple Mention Detection tool that surprisingly performs better than state-of-the-art systems
over the described data. % ns

% ns
The paper is organized as follows. We describe related benchmarks in Section \secref{sec:related} and the new 
benchmark in Section \secref{sec:benchmark}. Then in Section \secref{sec:experiments} we report the performance of a state-of-the-art system on the benchmark. 
We conclude with summary in Section \secref{sec:conclusions}.

%%% Local Variables: 
%%% mode: latex
%%% TeX-master: "graphq16-keyword-search"
%%% End: 

\section{Related Work}
\label{sec:related}
There are several annual Mention Detection challenges that publish benchmarks, such as the TAC-KBP 
(Text Analytic Conference - Knowledge Base Population
\cite{TAC-KBP})\footnote{https://tac.nist.gov//2016/KBP/}, the Microspots NEEL 
(Named Entity rEcognition and Linking)\footnote{http://microposts2016.seas.upenn.edu/challenge.html} and ERD (Entity Recognition and Disambiguation~\cite{ERD'14}).
In addition there are benchmarks published by 
specific research groups % ns individuals
such as AIDA~\cite{AIDA}, AQUAINT~\cite{AQUAINT}, MSNBC~\cite{ACE-MSNBC} and more. 

The Gerbil project~\cite{GERBIL} is a framework for the evaluation of Mention Detection tools. 
It defines formats and APIs for adding new benchmark data and new Mention Detection tools. 
The project contains $19$ datasets, among them the above mentioned AIDA and AQUAINT.

Most of those benchmarks focus on named entities 
% ns only as those entities are well defined. 
probably as those entities are well defined. 
Few benchmarks such as~\cite{AQUAINT,ACE-MSNBC} cover also general entities, but they are 
relatively small, containing only few hundred Mentions. 
Moreover, all existing benchmarks 
% ns are for written text and there are no Mention detection benchmarks on spoken data.
refer to written text only. 

% ns
% In this paper we tackle the above two shortcomings of existing Mention detection benchmarks. 
% First, our benchmark has both named and general entities and second, 
% they are annotated both on written as well as on noisy spoken data.
In contrast, the benchmark data described in this work covers both named entities and more general entities,
with respect to written Wikipedia text as well as relatively noisy spoken data. 

%%% Local Variables: 
%%% mode: latex
%%% TeX-master: "graphQ16-keyword-search"
%%% End: 

\section{The Benchmark}
\label{sec:benchmark}
The benchmark was built using 
% ns ten labelers on a crowd source site
the CrowdFlower platform.\footnote{\url{https://www.crowdflower.com/}}
% ns The site supports high quality workers through test questions that are planted inside jobs. 
This platform enables relying on high quality workers by integrating hidden test questions within the annotation task, 
and considering only the work done by annotators who correctly answered a pre--determined fraction of these. 
% ns The job owner can define the minimal passing grade and labelers that fail to pass are blocked from the job. 

For building the benchmark, we employed a two stage process -- a detection task, followed by a confirmation task. 
% ns First, a detection job and then a confirmation job.  

In the detection task, % ns job, 
labelers were presented with a text (usually a 
single % ns
sentence)
% ns to annotate and the task was to highlight terms from the text (where a term can be one or more tokens)  and
and were asked to link 
terms in the text % ns them 
to a Wikipedia page.\footnote{We use DBPedia and Wikipedia interchangeably 
as one is a reflection of the other}. 

%For example a valid annotation for the text given in~\ref{box:box2} is 
%\begin{verbatim}
%empire#uri:Empire
%Pedro II#uri:Pedro_II_of_Brazil
%\end{verbatim}

Then, in the confirmation 
task, the % ns job, 
labelers were presented with the text and the union of all Mentions 
identified in the % ns from the 
detection task, % ns job and the task was 
and were asked to confirm or reject each Mention.
%according to the task guidelines. % ns
%Each text was labeled by ten labelers in each of the tasks (i.e., detection and confirmation). 

To accommodate the issues of generality (what to annotate), nesting of Mentions, and specificity, 
we employ the following measures.  
First, each text was associated with a related 
general % ns 
{\it topic\/}.
% that is a Mention by itself.
The topics were selected at random from Debatabase\footnote{http://idebate.org/debatabase}.
%covering a wide variety of domains, from atheism to
%the role of wind power in future energy supply
For example, Text~\ref{box:box2} above 
was associated with the topic \texttt{We should abolish the Monarchy}. Overall, we selected $81$ topics. 
We then defined the following guidelines for the labelers.
\begin{enumerate}
\item Generality - General terms that clearly have no relation to the topic, should not be marked.
\item Nesting - The longest phrase that corresponds to a single Wikipedia title should be marked.
\item Specificity - The selected Wikipedia title should clearly match the meaning of the marked term, 
% ns based on the provided context.
in the context of the provided topic.
\end{enumerate}
The full guidelines of the two tasks can be found in Appendix~\secref{sec:guidelines} below.
Figures~\ref{fig:detection} and~\ref{fig:confirmation} show the User Interface of the detection and confirmation tasks. In the detection, the labelers 
were instructed to enter the detected Mentions into a field, each Mention in a separate line in the form of $<$term$>$\#$<$link$>$. 
\begin{figure}[tbh]
\center
  \includegraphics[width=3.3in]{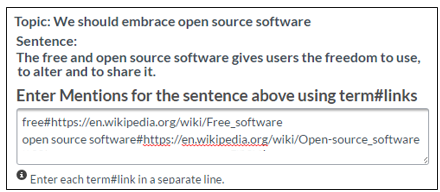}
  \caption{The detection UI}
  \label{fig:detection}
\end{figure} 

\begin{figure}[tbh]
\center
  \includegraphics[width=3.3in]{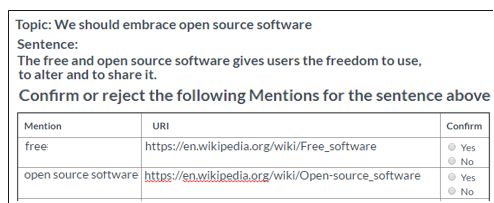}
  \caption{The confirmation UI}
  \label{fig:confirmation}
\end{figure} 

% ns In the confirmation task, 
Since the labelers were instructed to select all valid pages for a given term, % ns hence 
the ground truth can contain multiple correct pages for some of the terms (as long as they were confirmed by the majority of the annotators as described above). 
Furthermore, to have a full coverage of the text, the labelers were instructed in both jobs to detect and 
confirm all repeating occurrences of the same term. Thus, for a given text the same term can appear multiple times in Mentions (either with the same span but linked to different entities, or with different spans in the text).   

The ground truth was 
then defined % ns created by taking 
as Mentions that were confirmed by the majority of the annotators % ns voters 
(i.e., at least six out of ten) in the confirmation task. % ns

The full benchmark consists of $1000$ sentences from Wikipedia and $1000$ sentences of spoken % ns speech 
data.
The sentences were selected as follows.
For the Wikipedia sentences, we selected Wikipedia articles that discuss the above $81$ topics, and then we picked 
$1000$ sentences at random from those articles. We refer to this dataset by \Wiki. 

The spoken data sentences were taken 
from professional speakers
discussing % ns 
some of those topics. 
The spoken data has two forms: % ns flavors, one that was generated by 
the output of 
an Automatic Speech Recognition (ASR) engine;
 % ns and the second was manually transcribed and further cleansed 
and a cleansed manual transcription of it. The generation of the spoken data is described in~\cite{DEBATER-OSA}. 
The $1000$ sentences of the spoken data were selected at random from those speeches. 
We refer to the two flavors of the spoken data by \ASR~and \Trans, 
where the former is the output of the ASR engine without any editing, and the latter is the output of the manual transcription.

The \Trans~data is 
naturally % ns 
cleaner than the \ASR~data, and therefore
the labeling of the spoken data and the generation of the ground truth was done on \Trans. 
However, since for applications that work directly with speech, only ASR output will be available, 
%without cleaned manual transcripts, 
we projected the ground truth of \Trans~to \ASR, resulting in the \ASR~benchmark which enables the evaluation of models for Mention Detection in spoken data.

The projection of the ground truth of \Trans~to \ASR~was done as follows.
Given a pair of corresponding sentences $(T,A)$ where $T$ is from \Trans~and $A$ is from \ASR, we create a mapping from each offset (i.e., character) in $T$ to a corresponding offset (i.e., character) in $A$, by using minimum Edit distance with backtracking \cite{Hall:1980}. 
Then given a labeled Mention $(t,s,e)$ in $T$, we use the above mapping to find the corresponding span $s'$ in $A$
and define the corresponding Mention in $A$ as $(t',s',e)$, where $t'$ is the text in span $s'$ in $A$. 

An example of the projection is illustrated in Text~\ref{box:box3}.
% ns a sentence was labeled on the \Trans~data and the labeling was carried over to the \ASR. 
Note the distorted \term{jewish refute. these} that is linked to \entity{Jewish\_refugees}. 
The ASR data poses therefore an additional challenge for Mention Detection over the base task. 

\begin{story}[ht]
  \framedbox[0.4\textwidth]
		{\Trans: \mention{Jewish refugees} were turned away from the \mention{UK}\\
		 \ASR: \mention{jewish refute. these} were turned away from the \mention{u\_k} }
		\caption[text]{}
  \label{box:box3}
\end{story}

% that are closest to the actual data that will be used in real-time.

% and open a field of detecting Mentions from such noisy data. 
%For example the Trans text\\
 %
%\noindent 
%\fbox{
			%\parbox{1.0\linewidth} {%
							 %\mention{Jewish refugees} were turned away from the \mention{UK}%
							%}%
		 %}\\
%
%
%appears in the ASR dataset as \\
%
%\noindent 
%\fbox{
			%\parbox{1.0\linewidth} {%
							 %\mention{jewish refute. these} were turned away from the \mention{u\_k}%
							%}%
		 %} 
%

Overall, the \Wiki~dataset has $6375$ Mentions, out of which only $486$ are named entities and the rest are general entities.
Each of the \ASR~and \Trans~has $6239$ Mentions, out of which only $84$ are named entities.
The low number of named entities compared to general entities, is an evidence to the importance of Mention Detection benchmarks
such as the one described in this paper. 

Note that on average there are about $6.2$ Mentions per sentence in each of the datasets. 
Given an average sentence length of $20$ tokens, and given that some Mentions cover more than one token, this is quite a robust coverage of the text.

% ns We discuss now the agreement between labelers. 
The average pair-wise kappa\cite{Kappa} of the detection 
task % ns 
% ns was $0.3$ for Wiki and and $0.34$ for Trans. 
was $0.3$ for \Wiki~and and $0.34$ for \Trans. 
The average kappa for the confirmation 
task % ns 
% ns was $0.47$ for Wiki and  $0.54$ for Trans. 
was $0.47$ for \Wiki~and  $0.54$ for \Trans. 
Since different labelers labeled different number of sentences, the kappa for each pair of labelers was calculated by 
considering the sentences they both labeled. The average kappa over all pairs of labelers was taken as a weighted average of their kappa, 
where the weight of each pair was the total number of Mentions in their shared sentences. 

The relatively low kappa in the detection 
task % ns 
is attributed to the fact that this task is naturally more open ended compared to the confirmation task in which annotators
simply need to confirm or reject candidates out of a fixed list of candidate Mentions. 
Moreover, dealing with general entities adds to the inherent complexity of the detection task.
Note that the goal in the detection sub-task was to obtain a high coverage of Mentions whose union is then used in the confirmation sub-task. 
The low kappa in the detection, indicates a divergence between labelers and thus a high coverage of candidate Mentions.

In the confirmation sub-task, whose output determines the gold standard, the union of all detected mentions is shown to the labelers. 
%Note that on average there were more than six confirmed mentions per sentence which indicates on a high coverage. 
% in the detection task there was no 
% fixed set of Mentions and users had to mark as many Mentions as they see. 
% Some of them marked just one or two Mentions for each sentence. 
% This was enough for them to pass the test questions. 

The combination of low kappa in the detection task and the relatively high kappa in the confirmation task justifies the two tasks
and indicates on a high coverage (in the detection task) and a high quality of Mentions (in the confirmation task). 
%%%

%%%

% The kappa in the confirmation task was much higher due to two reasons. First, the confirmation task is easier 
% since users just have to confirm or reject Mentions. 
% Second, there was a fixed set of Mentions while in the detection task, labelers had to select their own Mentions. 
  	
% We plan to release the full benchmark if and after the paper is accepted. 
% ns --

The full benchmark is available for download at \url{http://www.research.ibm.com/haifa/dept/vst/debating_data.shtml}.

\section{Evaluation}
\label{sec:experiments}
As described in Section~\secref{sec:benchmark} above, our benchmark consists of $1000$ labeled sentences from 
Wikipedia,  $1000$ labeled sentences of 
transcribed (Trans) % ns Trans 
spoken data,
and $1000$ sentences of ASR data.  
 
We divided each of the three datasets to two equal parts of training and testing, each with $500$ sentences. 
We refer to them as Wiki-dev, Trans-dev and ASR-dev for development and Wiki-test, Trans-test and ASR-test for testing.

Table~\ref{tab:test-500} shows the results of a state-of-the-art system TagMe~\cite{TAGME} on the three test datasets. To accommodate guideline 2 (i.e., Nesting of Mentions, as described in Section~\ref{sec:benchmark}), we configured TagMe to return the longest phrases and avoid nesting of Mentions.

We can see that the precision and recall on \Wiki~are higher than on the two spoken datasets \Trans~and \ASR. For example the recall on \Wiki~is $0.523$ compared with $0.436$ on \Trans~and $0.421$ on \ASR. This is expected as the Wikipedia text is cleaner. We can also see that the performance on the \Trans~dataset 
is % ns is 
a bit higher than on the \ASR~dataset. This is also expected as the \ASR~data is more noisy.
\begin{table}[tbh]
\center
\small

\eat{
\begin{tabular}{||l|l|l|l|l||}
 \hline
  Dataset & Metric & TagMe  \\
 \hline
\multirow{3}{*}{Wiki}    		 & Precision &	0.584     \\
										 & Recall 	   &	0.523	  \\
                             				& F1  		   &	0.552 	  \\ \hline
\multirow{3}{*}{Trans}       & Precision &	0.569 	  \\ 
										& Recall 	   &	0.436 	  \\ 
                             				& F1      	 	   &	0.494 	  \\ \hline
\multirow{3}{*}{ASR}         & Precision  &	0.555 	  \\
										& Recall 	   &	0.421 	  \\
                             				& F1  		  &	0.478 	  \\ \hline
\end{tabular}
}

\begin{tabular}{||l|c|c|c||}
 \hline
  & Precision & Recall & F1-measure  \\
 \hline
 Wiki 		&	0.584 &  0.523 & 0.552   \\
 Trans		&  0.569 &  0.436 & 0.494   \\
 ASR			&  0.555 &  0.421 & 0.478  \\
 \hline 
\end{tabular}

\caption{Performance of TagMe on the test sets}
\label{tab:test-500}
\end{table}

%%% Local Variables: 
%%% mode: latex
%%% TeX-master: "graphq16-keyword-search"
%%% End: 

\section{Conclusions}
\label{sec:conclusions}
We presented a large scale Mention Detection benchmark that contains named entities as well as general entities, annotated on  
both clean, written text and noisy, speech data.  
The benchmark contains full coverage of annotations over $1000$ sentences from Wikipedia and $1000$ sentences of speech data that appear in two forms, one transcribed 
manually, and the other is the output of an ASR engine. % ns  with about $6500$ Mentions in each.
The benchmark was 
annotated via % ns labeled in 
a high quality and controlled crowd sourcing process, % with 
based on % ns 
clear guidelines % ns on 
indicating
what to annotate and how to resolve nesting and specificity of conflicting Mentions. 
Each of the datasets includes a total of around $6500$ Mentions, where the named entities are less than $8\%$ of them and the rest are general entities. 

The benchmark is robust in terms of the types of annotated Mentions and the coverage of the underling texts, and thus can be used for the evaluation of NLP applications that require semantic understanding of text. 

% The presented tool works better than state-of-the-art systems and handles the specific issues that 
% are raised by general Mentions.

%%% Local Variables: 
%%% mode: latex
%%% TeX-master: "graphq16-keyword-search"
%%% End: 

\section{Acknowledgments}
\label{sec:ack}
We would like to thank Dafna Sheinwald for comments on earlier drafts of this paper, to Michal Jacovi and to Alon Halfon for the help with the setup of the work on CrowdFlower, and to the anonymous labelers that made that benchmark available. 
% \nocite{*}
\section{Bibliographical References}
\label{main:ref}

\bibliographystyle{lrec}
\bibliography{references}
\section{Appendix: Labelers Guidelines}
\label{sec:guidelines}
We describe below the guidelines that were given to the labelers in the detection and in the confirmation tasks. 
\subsection{Detection Task}
\label{subsec:detection}
In this task, you are given a topic and a free text paragraph related to this topic. Your task is to detect terms mentioned in the text, and link those terms to their most relevant Wikipedia title, based on the following guidelines.

\begin{enumerate}
\item The longest phrase that corresponds to a single Wikipedia title should be marked. For example, if the text mentions the term \term{video games} you should mark this term along with the Wikipedia title \entity{Video\_game}, and not mark the individual terms \term{video} (with \entity{Video}) nor \term{games} (with \entity{Game}).
\item Terms should be entered into the text field, in separated rows in the same order as they appear in the original text. Each term should be entered as $<$term$>$\#$<$link$>$ where $<$term$>$ is the exact original phrase as appears in the text and the link is the URL of the corresponding Wikipedia title.
\item Derivations and/or Redirects should be considered and used. For example, the term \term{students} should be linked to its derivation in Wikipedia, \entity{Student}; the term \term{election campaign} should be linked to \entity{Political\_campaign} since it is redirected in Wikipedia to this title; and so on.
\item General terms that clearly have no relation to the topic, should not be marked. 
In particular, general terms that undoubtedly convey no content related to the 
pre-specified topic - e.g., \term{first}, \term{known}, \term{today}, \term{different},
\term{numbers}, etc., should not be marked, even though they may have a corresponding Wikipedia title. 
\item Disambiguation should be done based on context. 
If a term can be associated with several Wikipedia titles, you should link it with the title that best matches its meaning, based on the context of the entire text paragraph, 
and also - if needed - based on the pre-specified topic. 
For example, in a text discussing wild animals, the term \term{Jaguar} should be linked to
the Wikipedia title that describes this animal, 
and not to a Wikipedia title discussing Jaguar cars. 
\item The selected Wikipedia title should clearly match the meaning of the marked term. 
In case of a term that has both a general title and a more specific context-dependent title, the specific title should be selected. 
For example in a text that talks about Israel and mentions the term \term{air force}, the term should be linked to \entity{Israeli\_Air\_Force} and not to \entity{Air\_force}.
\end{enumerate}

\textbf{Technical Guidelines}
\begin{enumerate}
\item It is recommended to use the flexible search interface of Wikipedia to find the relevant Wikipedia title that matches the term. 
\item Names and titles should be marked together, if possible. For example the phrase \term{US Secretary of State Hillary Clinton} should be linked as a whole to  \entity{Hillary\_Rodham\_Clinton}
\item The selected Wikipedia title should not correspond to an internal Wikipedia section,   nor to a page of type list (such as \entity{List\_of\_political\_scientists}) and neither to a page of type disambiguation (such as \entity{Map\_(disambiguation)}).
\item Anaphora and co-references should not be resolved. In particular, pronouns like \term{he}, \term{they}, etc., should not be marked as terms.  
\item If a term appears several times in the text with the same meaning, and you decided to link this term to a particular Wikipedia title, then you should make sure to link all its occurrences in the text to the same title.
\end{enumerate}

\subsection{Confirmation Task}
\label{subsec:confirmation}
In this task you are given a topic, a free text sentence related to this topic, and a list of terms mentioned in this sentence, linked to their presumably corresponding Wikipedia titles.  Please confirm or reject each suggested term, according to the following guidelines.

\begin{enumerate}
\item The marked term should represent the longest phrase that corresponds to a single Wikipedia title. 
For example, if the term \term{video games} is linked to the Wikipedia title \entity{Video\_ game}, and its sub terms \term{video} and \term{games} are linked to Video and Game, respectively, then you should confirm \term{video games} and reject \term{video} and \term{games}.  
\item Derivations and/or Redirects should be considered and used. 
For example, the term \term{students} should be linked to its derivation in Wikipedia, \entity{Student}; 
the term \term{election campaign} should be linked to \entity{Political\_campaign} 
since it is redirected in Wikipedia to this title; and so on.
\item General terms that clearly have no relation to the topic, should not be confirmed. 
In particular, general terms that undoubtedly convey no content related to the pre-specified topic - e.g., \term{first}, \term{known}, \term{July}, \term{today}, \term{different}, \term{numbers}, etc., 
should not be marked, even though they may have a corresponding Wikipedia title. 
\item Disambiguation to a Wikipedia title should be done based on context. 
If selecting the Wikipedia title associated with a term involves disambiguation, this should be done based on the context of the surrounding text, and also - if needed - based on the pre-specified topic. 
\item The selected Wikipedia title should clearly match the meaning of the marked term. 
In case of a term that has both a general title and a more specific context-dependent title, the specific title should be selected. 
For example in a text that talks about Israel and mentions the term \term{air force}, the term should be linked to \entity{Israeli\_Air\_Force} and not to \entity{Air\_force}.
\end{enumerate}

\textbf{Technical guidelines.}
\begin{enumerate}
\item Names and titles should be marked together, if possible. For example the phrase \term{US Secretary of State Hillary Clinton} should be linked as a whole to  \entity{Hillary\_Rodham\_Clinton}.
\item The selected Wikipedia title should not correspond to an internal Wikipedia section,   nor to a page of type list (such as \entity{List\_of\_political\_scientists}) and neither to a page of type disambiguation (such as \entity{Map\_(disambiguation)}). 
\item Anaphora and co-references should not be resolved. In particular, pronouns like \term{he}, \term{they}, etc., should not be confirmed as terms.  
\item The same term may appear multiple times in a single sentence. 
For example in the text ``As fewer choices are offered to the voters, voters may vote for a candidate \ldots'', 
you may see two candidates for \term{voters}: voters(36,42) (where (36,42) is the span of \term{voters} in the text) and for voters(44,50).  
You should confirm/reject each appearance independently.
\item In addition, in principle a single term may be associated with several Wikipedia titles, as long as the guidelines above are satisfied. For example, in the text ``\ldots complained about African marriage customs \ldots'' the term \term{customs} may appear with a link to \entity{Convention\_(norm)}  and to \entity{Tradition}. Both options are valid and can be confirmed. 
\end{enumerate}

%\section{Language Resource References}
%\label{lr:ref}
%\bibliographystylelanguageresource{lrec}
%\bibliographylanguageresource{references}

\end{document}